\documentclass[letterpaper,twocolumn,10pt]{article}

\usepackage{mystyle}
\usepackage{epsfig,endnotes}
\usepackage{xcolor}
\usepackage{amssymb}
\usepackage{mathtools,algorithm,algpseudocode}
\usepackage[parfill]{parskip}
\begin{document}

\date{}

\title{\Large \bf Error-Aware B-PINNs: Improving Uncertainty Quantification in Bayesian Physics-Informed Neural Networks}

\author{
{\rm \hspace{0cm}}\\
\and
{\rm \normalsize \textbf{Olga Graf}}\\
{\rm \normalsize Technical University of Munich}\\
{\rm \normalsize \texttt{graf@ma.tum.de}}
\and
{\rm \normalsize \textbf{Pablo Flores}}\\
{\rm \normalsize Pontificia Universidad Cat\'{o}lica de Chile}\\
{\rm \normalsize \texttt{ptflores1@uc.cl}}
\and
{\rm \hspace{0cm}}\\
\and
\hspace{0.18cm}{\rm \normalsize \textbf{Pavlos Protopapas}}\\
\hspace{0.18cm}{\rm \normalsize Harvard University}\\
\hspace{0.18cm}{\rm \normalsize \texttt{pavlos@seas.harvard.edu}}
\and
\hspace{-0.02cm}{\rm \normalsize \textbf{Karim Pichara}}\\
\hspace{-0.02cm}{\rm \normalsize Pontificia Universidad Cat\'{o}lica de Chile}\\
\hspace{-0.02cm}{\rm \normalsize \texttt{kpb@ing.puc.cl}}
}

\maketitle

\begin{abstract}
Physics-Informed Neural Networks (PINNs) are gaining popularity as a method for solving differential equations. While being more feasible in some contexts than the classical numerical techniques, PINNs still lack credibility. A remedy for that can be found in Uncertainty Quantification (UQ) which is just beginning to emerge in the context of PINNs. Assessing how well the trained PINN complies with imposed differential equation is the key to tackling uncertainty, yet there is lack of comprehensive methodology for this task.  We propose a framework for UQ in Bayesian PINNs (B-PINNs) that incorporates the discrepancy between the B-PINN solution and the unknown true solution. We exploit recent results on error bounds for PINNs on linear dynamical systems and demonstrate the predictive uncertainty on a class of linear ODEs.
\end{abstract}

\section{Introduction}
Physics-Informed Neural Networks (PINNs) are deep neural networks (DNNs) that are able to encode differential equations (DEs) as a component of the neural network itself. Early results by Lagaris [Lagaris et al., 1998] who first proposed solving differential equations using neural networks were followed by a rapid growth of publications on PINNs in the recent years [Raissi et al., 2019; Hagge et al., 2017; Mattheakis et al., 2019; Mattheakis et al., 2020, Han et al., 2018; Raissi, 2018]. PINNs have many advantages as compared to traditional numerical solvers, e.g., they provide closed-form solutions, are mesh-free. i.e., enable on-demand solution computation after training, can discover new solutions fast by leveraging transfer learning, and many more. 

However, in order to truly flourish and get deployed in safety-critical applications, deep learning techniques like PINNs need to be \emph{reliable}, i.e., endowed with high-quality uncertainty quantification (UQ) methods. In the context of PINNs, UQ methods are still scarcely utilized, and there is no agreed gold standard. Some recent works on the topic include [Psaros et al., 2022; Zou et al., 2022; Zhang et al., 2018; Graf et al., 2021].
Recently, physics-informed Gaussian processes (PI-GP) [Raissi et al., 2017] and Bayesian PINNs (B-PINNs) [Yang et al., 2021] have been proposed as a counterpart to the classical Bayesian probabilistic machine learning techniques. An alternative line of research considers the use of generative adversarial networks where the Bayesian inference is performed in the latent space instead of the parameter space [Yang et al., 2018; Meng et al., 2021].

UQ in B-PINNs can be seen as more challenging than UQ in traditional numerical solvers due to multiple sources of uncertainty, ranging from noisy data to model architecture related issues. There is also a black-box uncertainty related to how well the network is actually informed about the governing physics equations. It stems from unsupervised nature of PINNs, but is often being obscured by considering noisy measurement data. As a rule, all the forementioned sources of uncertainty are modeled using one "catch-it-all" term. In this work, we set our goal to disentangle these uncertainties and show how augmenting separate terms in total uncertainty with useful error estimates can improve the predictive quality of UQ methods.

Our contributions can be summarized as follows:
\begin{itemize}
	\item[$\circ$] We empirically and theoretically analyze the aleatoric and the epistemic part of the total predictive variance in unsupervised B-PINNs. We conclude that existing B-PINN framework lacks the necessary component to incorporate the equation-related uncertainty.
	\item[$\circ$] We introduce the concept of \emph{Error-Aware B-PINNs} and consider \emph{pseudo-aleatoric uncertainty} in the predictive variance.
	\item[$\circ$] We corroborate our concept in practice on a class of linear dynamical systems, for which we are able to rigorously model the pseudo-aleatoric uncertainty.
\end{itemize}

\section{Background}
\subsection{Problem Formulation}
PINNs can solve differential equations expressed, in the most general form, as follows (cf. [Psaros et al., 2022]):
\begin{align}\label{DE1}
	\mathcal{F}_{\lambda}[u(x)]=f(x),\quad  &x\in\Omega,\\
	\mathcal{B}_{\lambda}[u(x)]=b(x),\quad  &x\in\Gamma,\label{DE2}
\end{align} 
where $x$ is a space-time coordinate in a domain $\Omega$, $\mathcal{F}_{\lambda}$ is a general differential operator, $\mathcal{B}_{\lambda}$ is an initial/boundary condition (I/BC) operator acting on the domain boundary $\Gamma$, $f(x)$ is a source term, $b(x)$ is an I/BC term, and $\lambda(x)$ are the parameters related to the physics. Finally, $u(x)$ is the sought solution of the differential equation.

Various problem scenarios for \eqref{DE1}-\eqref{DE2} have been considered in the literature: $\mathcal{F}_{\lambda}$ and $\mathcal{B}_{\lambda}$ can be either stochastic or deterministic, known or unknown; $f$, $b$, and $\lambda$ can be either exactly known or partially known from noisy samples; solution $u$ can be either completely unknown or partially known from noisy data.

In this paper, we will focus on \emph{data-free deterministic scenario} with known and deterministic $\mathcal{F}_{\lambda}$, $\mathcal{B}_{\lambda}$, $f$, $b$, and $\lambda$. As discussed in the Introduction, this will help us to disentangle uncertainties arising due to noisy data and/or stochastic realizations from uncertainties arising due to non-ideal compliance with the underlying physics.

\subsection{Physics-Informed Neural Networks}
To solve differential equation \eqref{DE1}, a DNN, denoted by $u_{\theta}(x)$, is constructed as an approximation to the true solution $u(x)$. We use $\theta$ to denote all the DNN parameters that will be optimized during training. We will refer to this network as the \emph{surrogate network}. It is required that $u_{\theta}(x)$ complies with the physics imposed by the differential equation. Therefore, we define a \emph{residual network},
\begin{equation}\label{res}
	r_{\theta}(x):=\mathcal{F}_{\lambda}[u_{\theta}(x)]-f(x),
\end{equation} 
which shares the same parameters with network $u_{\theta}(x)$ and should ideally output zero for any input $x\in\Omega$. Network $r_{\theta}(x)$ is an example of a \emph{Physics-Informed Neural Network}.

There are two main ways of constructing the optimization objective, depending on how I/BCs are satisfied. So called soft I/BC enforcement incorporates I/BCs as a penalty term in the loss function. Hard I/BC enforcement encodes I/BCs in the network design and thus guarantees that they are exactly satisfied. In this paper, we resort to the latter approach and use a slightly modified version of Lagaris' reparameterization technique. E.g., in one-dimensional case, given an initial condition $u_0=u(x_0)$, we consider a transformation $\tilde u_{\theta}(x)=u_0 + (1-e^{-(x-x_0)}) u_{\theta}(x)$. With a slight abuse of notation, hereinafter we refer to $u_{\theta}(x)$ as the network which has already been reparameterized.

Mean square error (MSE) minimization is widely used in connection with PINNs, and in case of hard I/BC enforcement it simplifies down to minimization of the squared residual $r_{\theta}^2(x)$. Algorithm \ref{algo:1} summarizes our workflow of solving a differential equation with unsupervised PINN.

\begin{algorithm}
	\caption{Unsupervised PINNs for solving DEs}
	\label{algo:1}
	\begin{algorithmic}\small
		\State \textbf{Input:} $\mathcal{F}_{\lambda}$, $\mathcal{B}_{\lambda}$, $f(x)$, $b(x)$, $\lambda(x)$
		
        1) Construct $u_{\theta}(x)$ with random initialization of $\theta$.
        
        2) Transform $u_{\theta}(x)$ to encode $\mathcal{B}_{\lambda}[u_{\theta}((x)]=b(x)$.
		
		3) Construct $r_{\theta}(x):=\mathcal{F}_{\lambda}[u_{\theta}(x)]-f(x)$.
		
		4) Train with loss $r_{\theta}^2(x)$ to obtain point estimates $\hat \theta$.
		
		\State \textbf{Output:} $u_{\hat\theta}(x)$, $\forall x\in\Omega$
		
	\end{algorithmic}
\end{algorithm}

\subsection{Uncertainty Quantification in Bayesian Neural Networks}
Employment of \emph{Bayesian framework} for UQ in deep learning has proven very successful [Neal, 1993; MacKay, 1995; Lampinen et al., 2001; Graves, 2011]. Below we briefly review some concepts and methods that will be used throughout the paper.

Suppose we have a finite set of observations $\mathcal{O}$ on a bounded domain $\Omega_{\mathcal{O}}\subset\Omega$. Assume that observations are i.i.d. and drawn from a distribution $p(\mathcal{O}|\mathbf{\theta})$. This distribution is referred to as \emph{likelihood function} and is often modeled by a factorizable Gaussian distribution,
\begin{equation}\label{likelihood}
	p(\mathcal{O}| \mathbf{\theta})=\mathcal{N}(\mathcal{O};y_{\theta},\Sigma_{\mathcal{O}}),
\end{equation} 
where the mean is given by a relevant neural network $y_{\theta}$ and $\Sigma_{\mathcal{O}}=\sigma_{\mathcal{O}}^2 I$. It is therefore assumed that observations $\mathcal{O}$ are produced by a data-generating process which contains deterministic part modeled by $y_{\theta}$ as well as some additive noise. This noise represents irreducible or \emph{aleatoric uncertainty} and its covariance matrix $\Sigma_{\mathcal{O}}$ may be either known or assumed, or inferred from data.

Standard NN training is performed in a frequentist manner and does not account for uncertainty. The model parameters $\theta$ are inferred as point estimates by maximizing the likelihood, which, given that the likelihood is Gaussian, is equivalent to MSE minimization. Note that in case of unsupervised PINNs, in order to derive MSE loss as in Algorithm 1, we assume virtual "observations" which are always equal to zero, whereas the mean is given by a residual network $r_{\theta}$ and $\sigma_{\mathcal{O}}^2$ can be arbitrary. 

In contrast to frequentist inference, Bayesian inference places a distribution not only over the observations $\mathcal{O}$, but over the parameters $\theta$ as well, resulting in Bayesian neural networks (BNNs). The probabilistic nature of parameters in Bayesian framework gives rise to the \emph{epistemic uncertainty}.

The \emph{posterior distribution} over the parameters $\theta$, i.e., distribution over the "plausible" parameter values based on the observations $\mathcal{O}$, is given by Bayes' theorem, 
\begin{equation}\label{bayes}
p(\mathbf{\theta} | \mathcal{O}) = \frac{p(\mathcal{O} | \mathbf{\theta}) p(\mathbf{\theta})} {p(\mathcal{O})}, 
\end{equation} 
where $p(\mathbf{\theta})$ is the prior distribution over the parameters, and $p(\mathcal{O})$ is the evidence. Obtaining the posterior exactly via  \eqref{bayes} is often computationally and analytically intractable. A brief overview of techniques that can approximate the posterior by obtaining samples from it or by another distribution is provided in Section \ref{methods}.

The goal of UQ is to estimate the \emph{predictive distribution} $p(\hat y|x,\mathcal{O})$. Here, $\hat y$ denotes the prediction at $x$, and is endowed with a conditional distribution on given observations $\mathcal{O}$. It can be obtained by integrating out the model parameters $\theta$, or, if intractable, approximated using Monte Carlo estimation,
\begin{equation}\label{postpred}
p(\hat y|x,\mathcal{O})=\int p(\hat y|x,\theta)p(\theta|\mathcal{O})d\theta\approx \frac{1}{N}\sum_{i=1}^{N}p(\hat y|x,\theta^{\mathsf{*}}_{i}),
\end{equation} 
where $\theta^{\mathsf{*}}_{i}\sim p(\mathbf{\theta} | \mathcal{O})$ are the posterior samples. 

The mean of predictive distribution can be approximately calculated as
\begin{equation}\label{postpred_mean}
	\mathbb{E}[\hat y|x,\mathcal{O}]\approx \frac{1}{N}\sum_{i=1}^{N}y_{\theta^{\mathsf{*}}_{i}}(x):=\tilde\mu_{\mathcal{O}}(x).
\end{equation}
The variance of predictive distribution can also be approximately calculated. Using the law of total variance and assuming that the distribution $p(\hat y|x,\theta)$ has the same mean and variance as \eqref{likelihood}, we have
\begin{align}\label{postpred_var}
	\mathrm{Var}(\hat y|x,\mathcal{O})&=\mathbb{E}_{\theta|\mathcal{O}}[\mathrm{Var}(\hat y|x,\theta)]+\mathrm{Var}_{\theta|\mathcal{O}}(\mathbb{E}[\hat y|x,\theta])\nonumber\\
	&=\mathbb{E}_{\theta|\mathcal{O}}[\Sigma_{\mathcal{O}}]+\mathrm{Var}_{\theta|\mathcal{O}}(y_{\theta}(x))\nonumber\\
	&\approx\hspace{-0.03cm} \sigma_{\mathcal{O}}^2\hspace{-0.05cm}+\hspace{-0.05cm}\frac{1}{N}\hspace{-0.07cm}\sum_{i=1}^{N}(y_{\theta^{\mathsf{*}}_{i}}(x)\hspace{-0.05cm}-\hspace{-0.05cm}\tilde\mu_{\mathcal{O}}(x))^2:=\widetilde\sigma_{\mathcal{O}}^2(x).
\end{align}
The first and the second summand in \eqref{postpred_var} represent the aleatoric and the epistemic parts of the total uncertainty, respectively.

\subsubsection{Methods for Finding the Posterior Distribution}\label{methods}
The most popular posterior approximation methods can be divided into two families: sampling methods and variational methods. There is also an efficient closed-form solution for the simplified BNN, namely, if we discard parameter distributions in the hidden layers and keep them only in the last layer. We will refer to it as Neural Linear Model (NLM).

\textbf{Sampling Methods.} Markov-Chain Monte Carlo methods draw parameter samples from the posterior by using a Markov chain with $p(\theta|\mathcal{O})$ as its stationary distribution. Widely used methods are Hamiltonian Monte Carlo and Langevin Dynamics. The drawback of such methods is the computational cost. We will not consider them in this paper.

\textbf{Variational Methods.} Variational inference (VI) methods approximate $p(\theta|\mathcal{O})$ by employing a variational distribution $q_w(\theta)$ that is optimized with respect to $w$ by maximizing the evidence lower bound. Popular methods are Bayes by Backpropagation ([Blundell et al., 2015], utilizes factorized Gaussian variational distribution) and Monte Carlo Dropout ([Srivastava et al., 2014], samples are drawn by randomly dropping for each sample a fixed percentage of parameters).

\textbf{Exact Methods.} NLM [Snoek et al., 2015] represents compromise between tractability and expressiveness. One can interpret NLM as a learned feature basis $\Phi_w=[\phi_w(x_1),...,\phi_w(x_M)]^T$ on which Bayesian linear regression $\Phi_w\theta$ is performed. We will denote the residual network trained with NLM by $r^{\mathsf{NLM}}_{\theta}$ in order to differentiate from a standard BNN.

\begin{figure*}[h]
	%\vspace{.3in}
	\includegraphics[width=\linewidth]{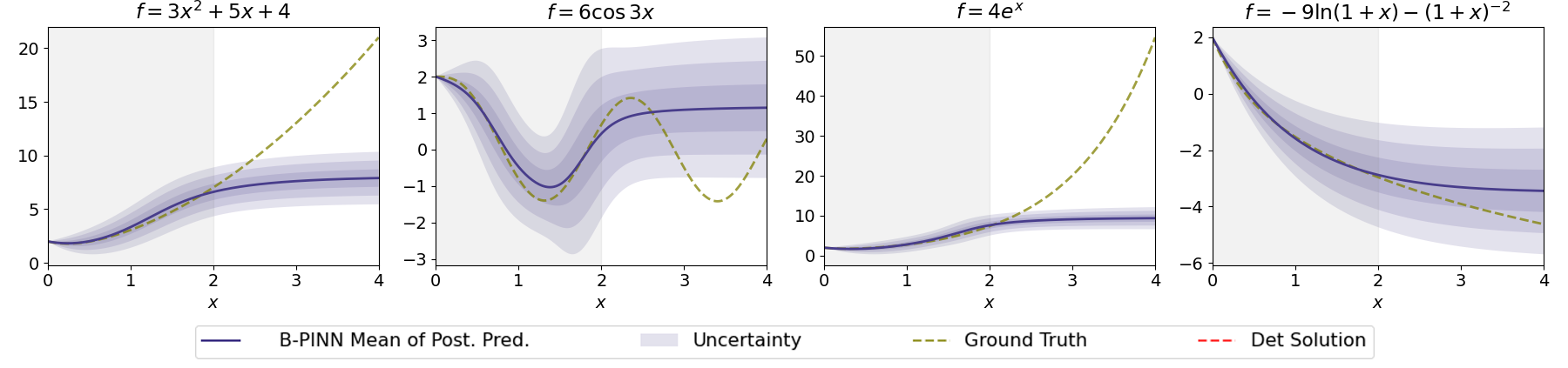}
	%\vspace{.3in}
	\caption{\textbf{Uncertainty estimation with existing B-PINN approach.} We solve the equation $u'+3u=f$ with various source terms, using methodolody from Section \ref{exist} combined with Variational Inference. We do not aim for the perfectly trained model on purpose. We see that uncertainty coming from underfitting is not always captured well, especially in the region where model was not trained (white region in the figures).}
	\label{fig1}
\end{figure*}

\section{Uncertainty Quantification in B-PINNs}
In the previous section, we have considered all observable information under one collective term $\mathcal{O}$. In fact, in the context of PINNs, these observations comprise a variety of sources: noisy and/or limited data, compliance with the governing physics equations, stochasticity in the physical model, neural network architecture. We will denote data-related observations by $\mathcal{D}$, equation-related observations by $\mathcal{P}$, and the remaining observations collectively by $\mathcal{H}$. As stated earlier, our primary goal is to study data-free deterministic case with a fixed DNN architecture, therefore we will focus on modeling the effect of $\mathcal{P}$ on uncertainty. Subsequently, it can be incorporated into the general framework, resulting in a predictive distribution $p(\hat u|x,\mathcal{D},\mathcal{P},\mathcal{H})$. Here, $\hat u$ denotes the probabilistic prediction for the DE solution.

To the best of our knowledge, equation-related observations for B-PINNs have not been thoroughly studied in the literature. One reason for that might be the fact that it is not possible to completely disentangle different sources of uncertainty:
\begin{itemize}
	\item[$\circ$] Observations of $f$ can be simultaneously seen as data-related observations $\mathcal{D}$ and 
	equation-related observations $\mathcal{P}$. Therefore, as it is not necessary to construct a special likelihood function for $\mathcal{P}$, this might obscure the importance of taking $\mathcal{P}$ explicitly into account at the later inference stages.
	\item[$\circ$] The total variance in \eqref{postpred_var} is affected by aleatoric uncertainty not only via the corresponding term, but also via the epistemic term, which is expected to increase with more noise. This can obscure the importance of properly modeling the aleatoric term $\sigma_{\mathcal{P}}^2$ in $\mathrm{Var}(\hat u|x,\mathcal{P})$. In fact, this term is absent in the existing B-PINN methodology, which might result in less reliable uncertainty estimates. 
\end{itemize}

\subsection{Existing Approach}\label{exist}
In this subsection, we will apply and analyze the original approach by Yang et al. [Yang et al., 2021] for B-PINNs in data-free case. We will empirically show that it does not always lead to useful uncertainty estimates, especially in the region where model was not trained.

Since we assume that $f$ is exactly known and there is no noise, we can consider a virtual dataset $\mathcal{D}=\{(x_j,0)\}_{j=1}^{M}$, whereas the corresponding likelihood is given by
\begin{equation}\label{likelihood_unsup}
	p(\mathcal{D}| \mathbf{\theta})=\prod_{j=1}^{M}p(0|x_j,\mathbf{\theta})=\prod_{j=1}^{M}\mathcal{N}(0;r_{\theta}(x_j),\sigma_{\mathcal{D}}^2).
\end{equation}%The usage of normal distribution in this case can be justified (see {\color{red}ref that Pavlos knows}). 
If $\sigma_{\mathcal{D}}^2$ is not known, it can be estimated, e.g., via hierarchical modeling. The prior distribution $p(\theta)$ is assumed to be Gaussian with zero mean and diagonal covariance matrix. Corresponding variances, if not known, can be inferred similarly to $\sigma_{\mathcal{D}}^2$.
Then, after we obtain samples from the posterior using methodology from Section \ref{methods}, we estimate the predictive distribution
\begin{equation}\label{postpred_unsup}
	p(\hat u|x,\mathcal{D})=\mathbb{E}_{\theta|\mathcal{D}}[\hat u|x,\theta].
\end{equation}
Since we are in data-free scenario, a natural choice for the distribution on the right-hand side of \eqref{postpred_unsup} is
\begin{equation}\label{deltadist}
	p(\hat u|x,\theta)=\delta(\hat u-u_{\theta}(x)).
\end{equation}
Substituting \eqref{deltadist} into \eqref{postpred_var} yields
\begin{align}\label{postpred_var_unsup}
	\mathrm{Var}(\hat u|x,\mathcal{D})\approx\widetilde\sigma_{\mathcal{D}}^2(x)=\frac{1}{N}\sum_{i=1}^{N}(u_{\theta^{\mathsf{*}}_{i}}(x)-\tilde\mu_{\mathcal{D}}(x))^2,
\end{align}
where only the epistemic term is present.

Although the predictive variance in \eqref{postpred_var_unsup} is "informed" about the underlying DE implicitly via $\mathcal{D}$, it might not be large enough to cover the true solution outside of the training region or when B-PINN is for some reason not trained well, as shown in Figure \ref{fig1}.

Another drawback of the current approach is its incompatibility with $r^{\mathsf{NLM}}_{\theta}$. By imposing $\mathcal{F}_{\lambda}[\Phi_w\theta]$, we lose the ability to derive a closed-form solution.

\subsection{Error-Aware B-PINNs}\label{our}
In classical function approximation problem with BNN, there is no guarantee that uncertainty covers the error $y-y_{\theta}$ in the region where no data is available, unless there are no specific assumptions on the function class. Unlike BNNs, B-PINNs are to a certain extent aware of discrepancy between the surrogate network and the true solution. Of course, this depends on the choice of the loss function as well as on the link between the chosen loss function and the error that the surrogate network makes.

In this subsection, we introduce a \emph{pseudo-aleatoric uncertainty} that quantifies the discrepancy between $u$ and $u_{\theta}$. We still call it aleatoric because it acts as a counterpart of aleatoric uncertainty in the classical Bayesian framework. On the other hand, it is not coming from data, but from PINN itself.

Our goal is to construct a distribution $p(\hat u|x,\theta,\mathcal{P})$ with the mean given by $u_{\theta}(x)$ and the pseudo-aleatoric variance given by some operator $\mathcal{E}$ which acts on pre-trained PINN,
\begin{equation}\label{cov_P}
	\sigma_{\mathcal{P}}(x)=\mathcal{E}[r_{\mathsf{MSE}}(x),\lambda(x)],
\end{equation}
where $r_{\mathsf{MSE}}$ denotes PINN trained with MSE. Clearly, this is not a straightforward task which raises questions how to construct $\mathcal{E}$ and how to find an appropriate functional form of the PDF. At this point, we are able to answer the first question with provable guarantees in the special case of linear dynamical systems (see Section \ref{example}). As for functional form of $p(\hat u|x,\theta,\mathcal{P})$, we consider normal distribution as a proof of concept. Although it does not represent the probability of the true solution, it makes sure that the uncertainty is inflated in the region where the error is high and that the true solution is covered by uncertainty. 

The predictive variance in the error-aware case is given by
\begin{align}\label{postpred_var_error}
	&\mathrm{Var}(\hat u|x,\mathcal{D},\mathcal{P})\nonumber\\
	&\approx\widetilde\sigma_{\mathcal{P}\mathcal{D}}^2(x)=\sigma_{\mathcal{P}}^2(x)+\frac{1}{N}\sum_{i=1}^{N}(u_{\theta^{\mathsf{*}}_{i}}(x)-\tilde\mu_{\mathcal{D}}(x))^2.
\end{align}
In case of DEs that do not fall into category of Section \ref{example}, there is no rigorous theory on how to construct \eqref{cov_P}. Various heuristic approaches can be considered, whereas the simplest one is equating $\mathcal{E}[r_{\mathsf{MSE}}(x)] = r_{\mathsf{MSE}}(x)$ or $\mathcal{E}[r_{\mathsf{MSE}}(x_n)] = \sum_{i=0}^{n} r_{\mathsf{MSE}}(x_i)$. We test the latter in Section \ref{pde} on an example of a non-linear PDE.

Now we show how to combine our approach with $r^{\mathsf{NLM}}_{\theta}$ such that the closed-form solution still exists. Namely, our distribution with pseudo-aleatoric variance can be used as a likelihood function for simulated data $\mathcal{D}_{\mathcal{P}}=\{(x_j,u_{\mathsf{MSE}}(x_j))\}_{j=1}^{M}$, where $u_{\mathsf{MSE}}$ denotes surrogate network trained with MSE,  resulting in
\begin{align}\label{new_lik}
	p(\mathcal{D}_{\mathcal{P}}| \mathbf{\theta})
	&=\prod_{j=1}^{M}\mathcal{N}(u_{\mathsf{MSE}}(x_j);u_{\theta}(x_j),	\sigma_{\mathcal{P}}^2(x_j)).
\end{align}
If used in conjunction with this likelihood, $r^{\mathsf{NLM}}_{\theta}$ provides tractable inference. Note that $\sigma_{\mathcal{P}}^2(x_j)$ is now heteroscedastic. As long as Bayesian linear regression model with heteroscedastic variance is less commonly used, we give here an explicit form of the predictive distribution,
\begin{align}\label{NLM_err}
	p(\hat u_{\mathsf{NLM}}|x,\mathcal{D}_{\mathcal{P}})&=\mathcal{N}(\hat u_{\mathsf{NLM}};\mu_{\mathcal{ D}_{\mathcal{P}}}(x),\sigma^2_{\mathcal{D}_{\mathcal{P}}}(x))\\
	&=\mathcal{N}(\hat u_{\mathsf{NLM}};\Phi_w\mu_{\mathsf{p}},\sigma^2_{\mathcal{P}}(x)+\Phi_w\Sigma_{\mathsf{p}}\Phi_w^T),\nonumber
\end{align}
where posterior covariance matrix and mean are given by
\begin{align}\label{NLM_err2}
	\Sigma_{\mathsf{p}}&=(\Phi_w^T\Sigma_{\mathcal{P}}\Phi_w+\sigma^{-2}I)^{-1},\\
	\mu_{\mathsf{p}}&=\Sigma_{\mathsf{p}}(\Sigma_{\mathcal{P}}^{-1}\Phi_w)^Tu_{\mathsf{MSE}}(x_{\mathcal{D}}),
\end{align}
$x_{\mathcal{D}}=[x_1,...,x_M]^T$, $\Sigma_{\mathcal{P}}=\mathrm{diag}([\sigma_{\mathcal{P}}^2(x_1),...,\sigma_{\mathcal{P}}^2(x_M)])$, and $\sigma^2$ denotes prior variance.

Moreover, this setup for $r^{\mathsf{NLM}}_{\theta}$ allows for prior optimization. We require the predictive uncertainty to cover the error bounds given by $\sigma_{\mathcal{P}}(x)$, i.e., 
\begin{align}\label{NLM_priorb}
|\mu_{\mathcal{D}_{\mathcal{P}}}(x)-u_{\mathsf{MSE}}(x)|\leqslant 3\sigma_{\mathcal{D}_{\mathcal{P}}}(x)-\sigma_{\mathcal{P}}(x). 
\end{align}
We only consider priors that satisfy the above inequality. To find $\sigma$ that makes the posterior predictive follow the error bounds as closely as possible, we solve the following minimization problem,
\begin{align}\label{NLM_prioropt}
	\underset{\sigma}{\min}\|\mu_{\mathcal{D}_{\mathcal{P}}}-u_{\mathsf{MSE}}\|+ \|\sigma_{\mathcal{D}_{\mathcal{P}}}-\sigma_{\mathcal{P}}\|. 
\end{align}
This simple approach can be regarded as error visualization technique. It can be implemented on top of deterministically trained network at almost no computational cost, yet provides quick and reasonable inference for the goodness of fit. We show the results for $r^{\mathsf{NLM}}_{\theta}$ with optimal prior in Figure 2.

\begin{figure*}%[h]
	\includegraphics[width=\linewidth]{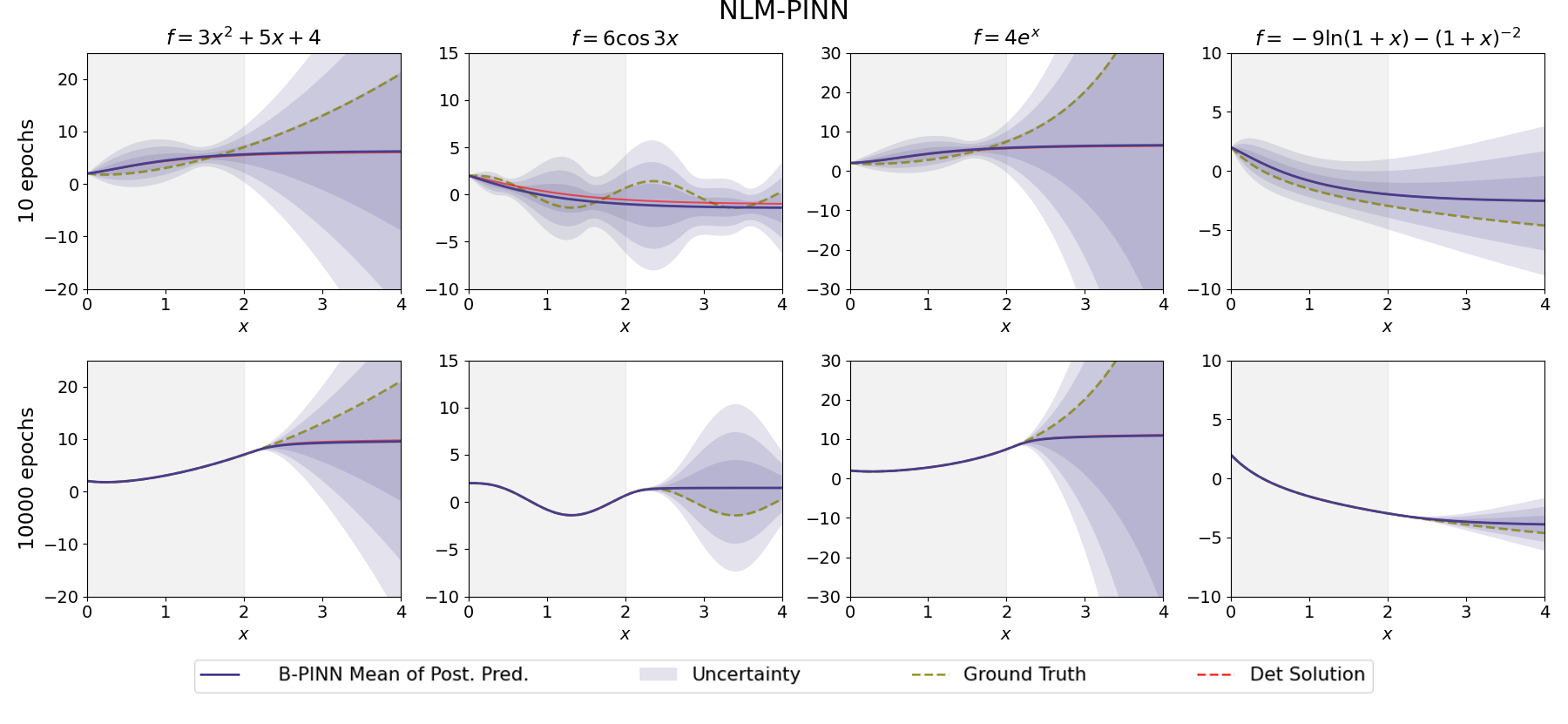}
	\includegraphics[width=\linewidth]{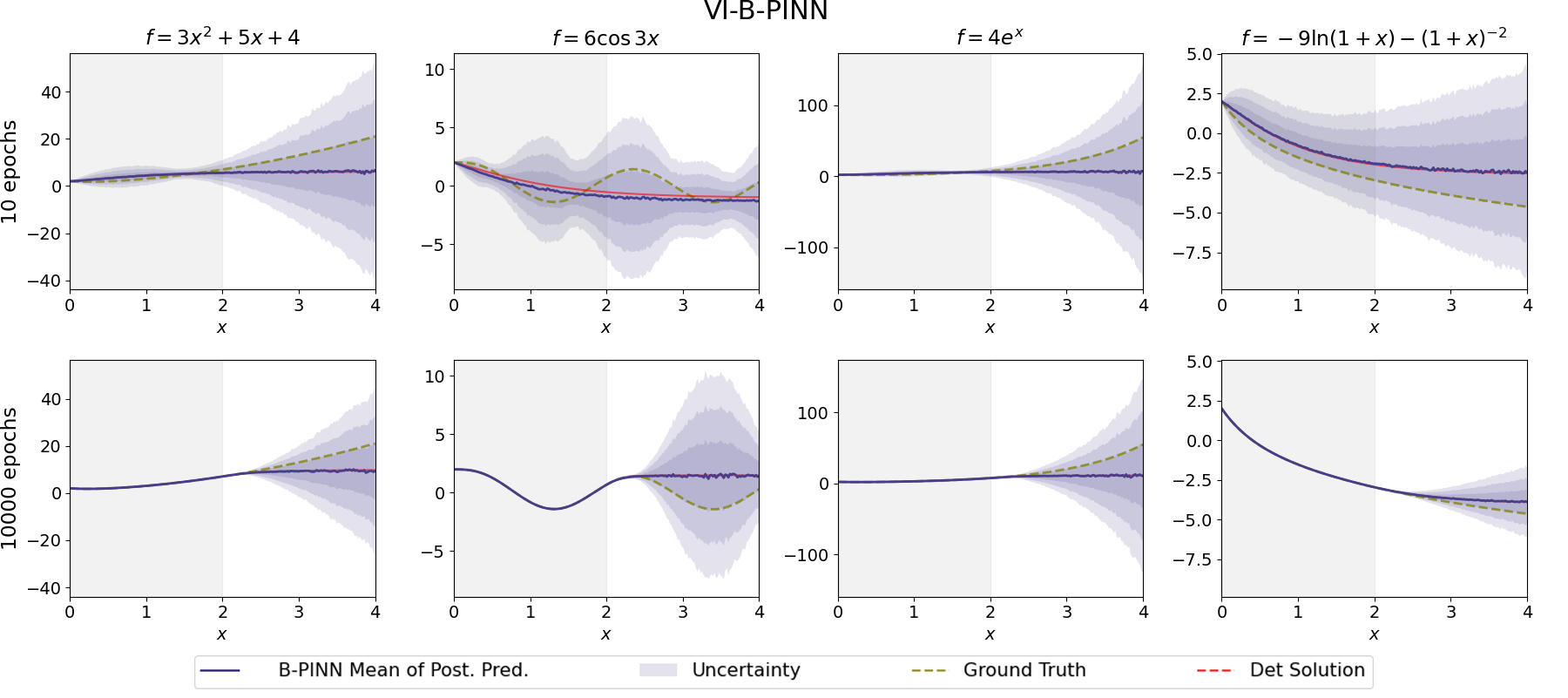}
	\caption{\textbf{Uncertainty estimation with Error-Aware B-PINNs.} We solve the equation $u'+3u=f$ with various source terms, using methodolody from Section \ref{our} combined with Variational Inference (VI-B-PINN) and Neural Linear Model (NLM-PINN). Our approach shows reasonable uncertainty estimates even in the test region where the model was not trained (white region in the figures) and in case it was trained only for 10 epochs.}
	\label{fig2}
\end{figure*}

\begin{figure*}%[h]
	\includegraphics[width=\linewidth]{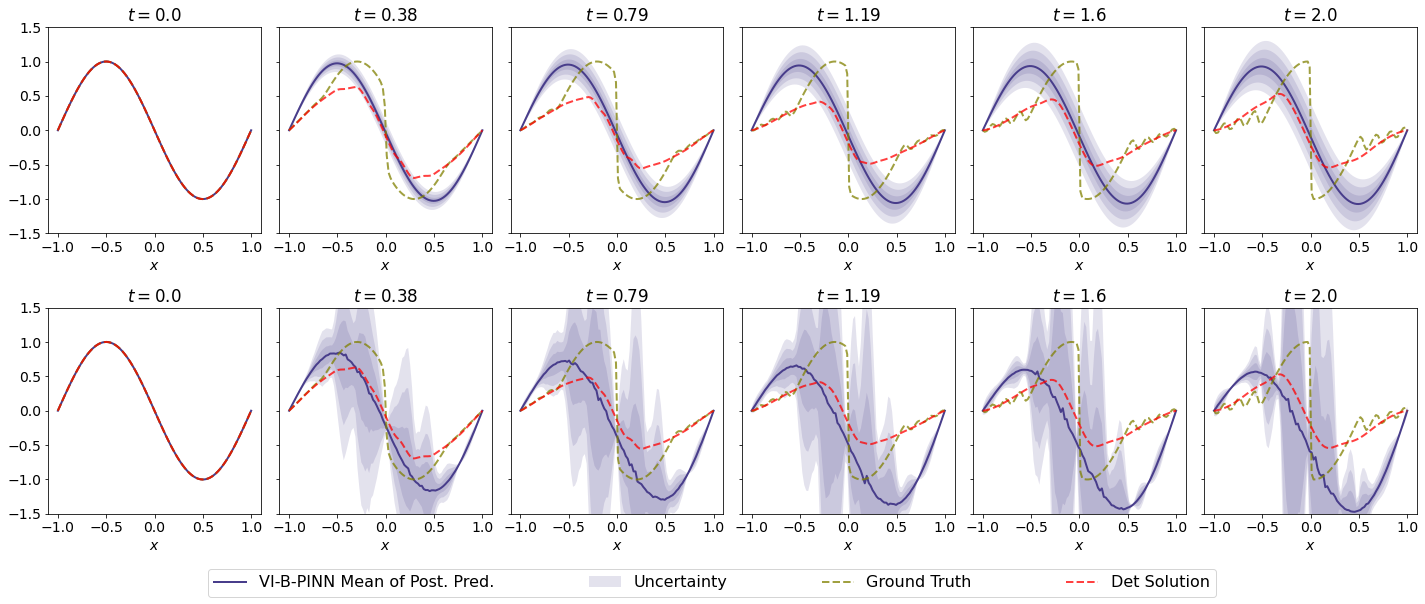}
	\caption{\textbf{Uncertainty estimation for Burgers’ equation.} We compare the existing B-PINN method (upper row) with our method (lower row). Uncertainty coming from underfitting is also not captured well due to our heuristic approach, nevertheless, provides more reasonable estimates than in the first case.}
	\label{fig:burgers}
\end{figure*}

Finally, we summarize our method for Error-Aware B-PINNs in Algorithm \ref{algo:2}.

\begin{algorithm}%[!t]
	\caption{Error-Aware B-PINNs for solving DEs}
	\label{algo:2}
	\begin{algorithmic}\small
		\State \textbf{Input:} $\mathcal{F}_{\lambda}$, $\mathcal{B}_{\lambda}$, $f(x)$, $b(x)$, $\lambda(x)$
		
        1) Construct and train $r_{\mathsf{MSE}}(x)$ as in Algorithm 1.
        
        2) Use $r_{\mathsf{MSE}}(x)$ to construct pseudo-aleatoric 
        
        uncertainty term $\sigma_{\mathcal{P}}(x)$.
        
        3) Construct Bayesian model $u_{\theta}(x)$ with $p(\theta)$.
        
        4) Transform $u_{\theta}(x)$ to encode $\mathcal{B}_{\lambda}[u_{\theta}((x)]=b(x)$.
		
		5) Construct B-PINN $r_{\theta}(x):=\mathcal{F}_{\lambda}[u_{\theta}(x)]-f(x)$.
		
		6) Obtain the posterior distribution of $\theta$ via one of the 
		
		techniques from Section $\ref{methods}$. Likelihood \eqref{likelihood_unsup} or \eqref{new_lik} 
		
		can be used for this purpose.
		
		7) Estimate mean and variance of $p(\hat u|x,\mathcal{D},\mathcal{P})$ via \eqref{postpred_var_error}.
		
		\State \textbf{Output:} Estimated $p(\hat u|x,\mathcal{D},\mathcal{P})$, $\forall x\in\Omega$
	\end{algorithmic}
\end{algorithm}

\subsubsection{Motivating Example: Error-Aware B-PINNs for Linear Dynamical Systems}\label{example}

One way to encode discrepancy between the surrogate network and the true solution into the distribution of $\hat u$ is to encorporate error bounds $|u(x)-u_{\theta}(x)|$ in its variance. We emphasize that while this does \emph{not} give the probability of seeing the true solution, it is a good starting point to show the importance of the aleatoric term in the total variance.

Recently, there appeared several works that study the failure modes and absolute error of PINN solutions [Liu et al., 2021; Graf et al, 2021]. In this subsection, we employ the approach of Liu et al. [Liu et al., 2021], which considers unsupervised scenario with hard I/BCs and derives error bounds which only depend on $x$ and $r_{\theta}$. It makes no assumption on the network architecture or whether the network is suffciently trained.

E.g., for the first order linear ODE, 
\begin{align}\label{1ode}
	u'(x)+\lambda u(x)=f(x), \quad \lambda>0,\quad u(x_0)\neq0,
\end{align}
the error bound is given by
\begin{align}\label{bound}
	|u(x)-u_{\theta}(x)|\leqslant e^{-\lambda x} \int_{x_0}^{x}e^{\lambda\xi} |r_{\theta}(\xi)|d\xi,
\end{align}
provided that $|r_{\theta}|$ is bounded on $\Omega$. We note that in practice we approximate this integral by summation, as we cannot evaluate the residual at every point in $\mathbb{R}^n$.

Inspired by this bound, we take
\begin{equation}\label{bnd}
	\sigma_{\mathcal{P}}(x)=\mathcal{E}[r_{\mathsf{MSE}}(x),\lambda]=e^{-\lambda x} \int_{x_0}^{x}e^{\lambda\xi} |r_{\mathsf{MSE}}(\xi)|d\xi.
\end{equation}

Assuming normal distribution, we obtain the results for several source terms and two inference strategies, Variational Inference (VI-B-PINN) and Neural Linear Model (NLM-PINN), as shown in Figure \ref{fig2}. 

Existing methodolody allows to consider also higher order linear ODE with constant and non-constant $\lambda(x)$, and systems of first order linear ODEs with constant coefficients. We present the results for the second order ODEs in Appendix.

\subsubsection{Outlook: Error-Aware B-PINNs for PDEs}\label{pde}
We consider Burgers' equation (non-linear PDE), 
\begin{equation}
\frac{\partial u}{\partial t}+u\frac{\partial u}{\partial x}=\nu\frac{\partial^2 u}{\partial x^2},
\end{equation}

as a proof of concept that our methodology might prove useful also in more advanced cases. The results are presented in Figure 3.

\section{Conclusion and Future Work}

Uncertainty quantification is a challenging task for deep learning models, especially if it is critical that the predictive uncertainty reflects the phenomena under consideration and is not giving misleading results. 

In this work, we tried to improve the UQ in Bayesian Physics-Informed Neural Networks by enforcing the information about network's compliance with the underlying differential equation. For that purpose, we introduced Error-Aware B-PINNs with the extra term dubbed as pseudo-aleatoric variance. We showed how to obtain this variance using the residual-based error bounds in case of linear dynamical systems. 

As a future extension, several non-linear ODEs can be considered by exploiting solution via perturbation series and constructing similar error bounds and error bound based variances. A more ambitious approach would be to establish distributions for the solution errors as opposed to the merely qualitative assessment of the presented approach.

\makeatletter
\renewcommand\@biblabel[1]{}
\makeatother
\nocite{*}
\bibliographystyle{apalike}
\bibliography{bibliography_short}

\end{document}

\date{}

\title{\Large \bf Supplementary Material:\\ Error-Aware B-PINNs: Improving Uncertainty Quantification in Bayesian Physics-Informed Neural Networks}

\maketitle

\begin{alphasection}
\section{Extended Results on Error-Aware B-PINNs for Linear Dynamical Systems}

\renewcommand{\thefigure}{4}
\begin{figure*}[h]
\centering
	\includegraphics[width=0.9\linewidth]{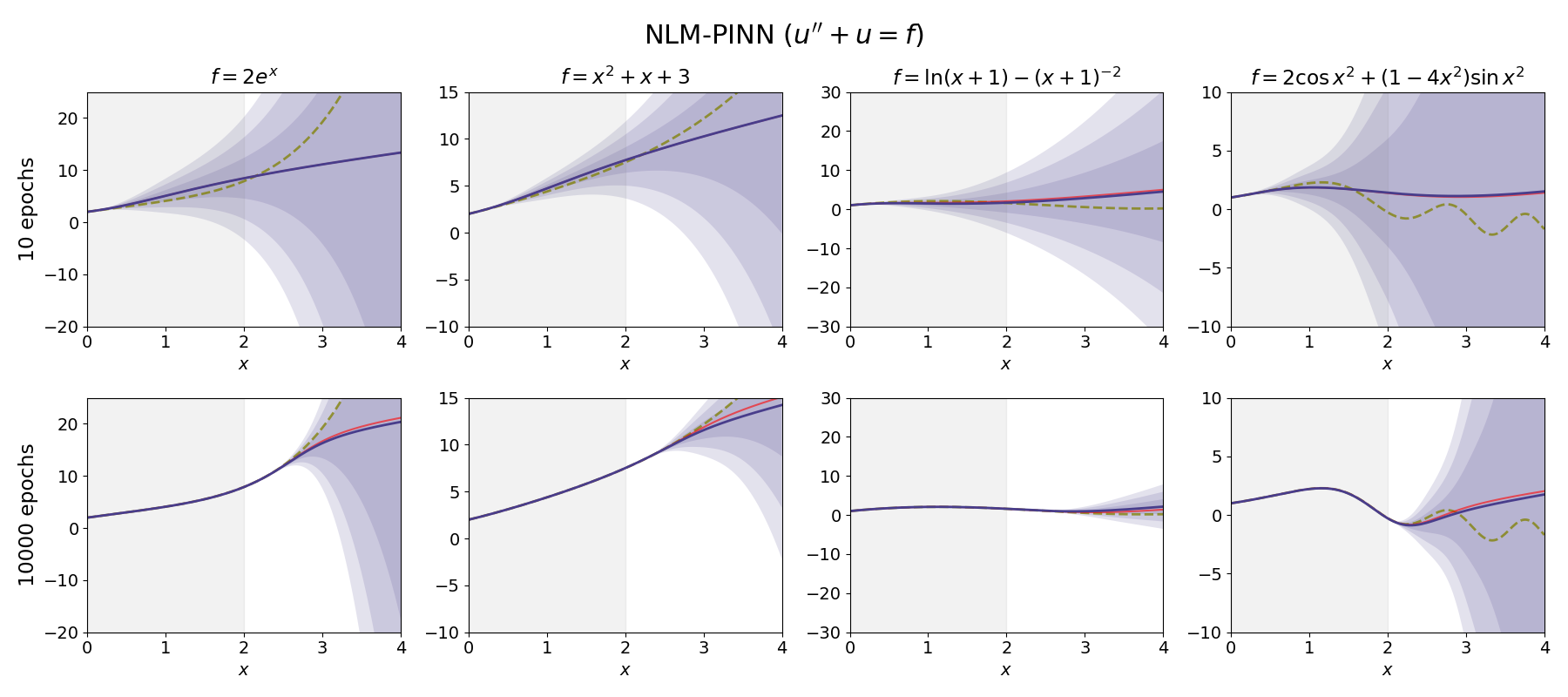}
	\includegraphics[width=0.9\linewidth]{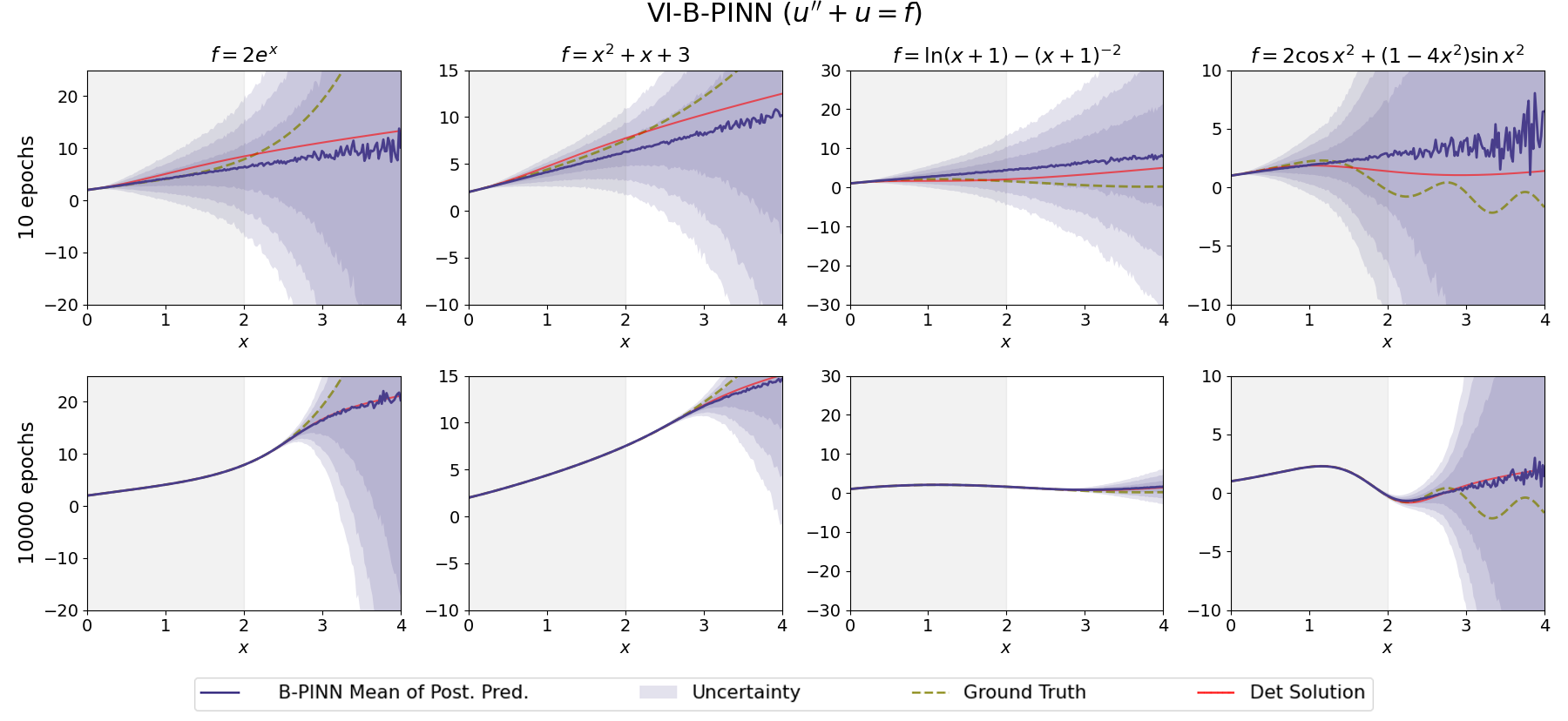}
	\caption{\textbf{Uncertainty estimation with Error-Aware B-PINNs.} We consider harmonic oscillator $u''+u=f$ under various source terms. We construct pseudo-aleatoric variance using error bounds, and test our method with Neural Linear Model (NLM-PINN) and Variational Inference (VI-B-PINN).}
	\label{secondorder1}
\end{figure*}

\thispagestyle{empty}

\newpage
\renewcommand{\thefigure}{5}
\begin{figure*}[h]
\centering
	\includegraphics[width=0.9\linewidth]{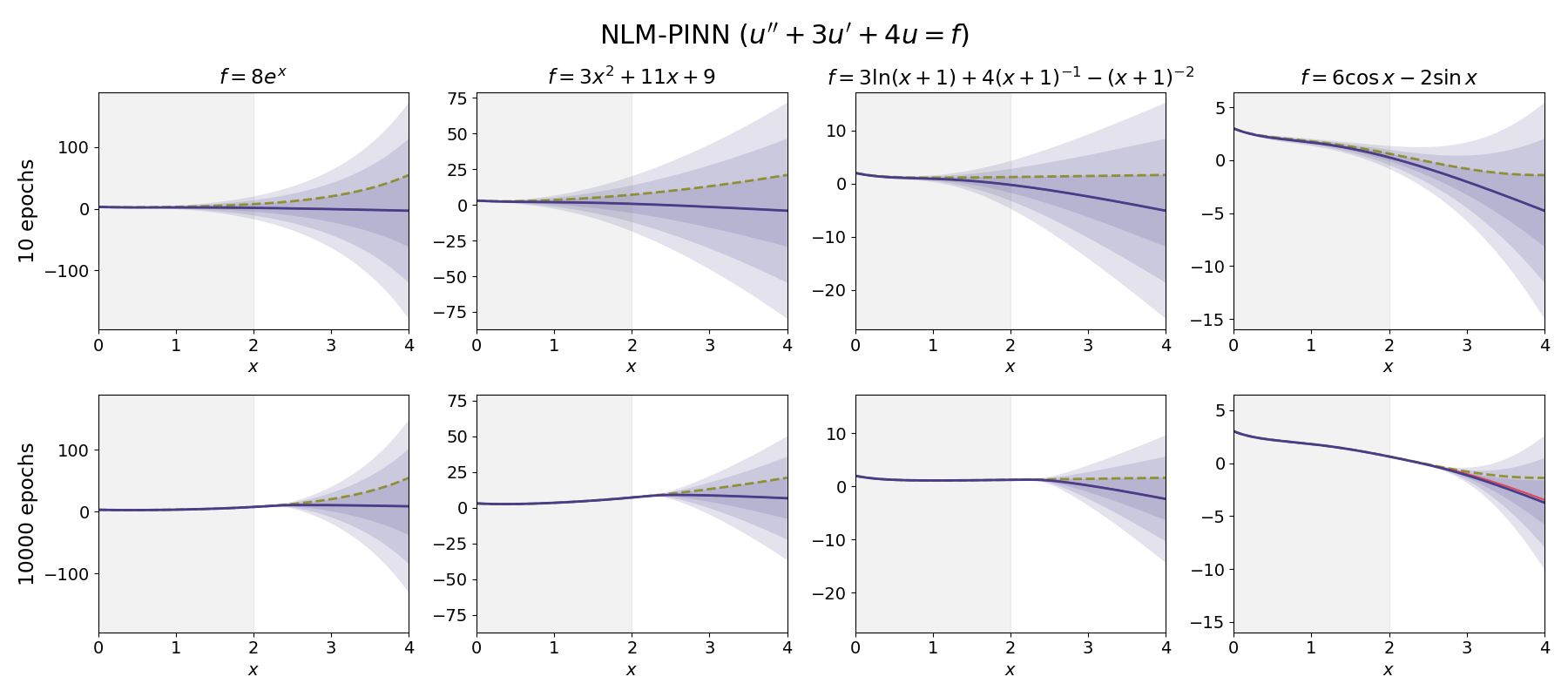}
	\includegraphics[width=0.9\linewidth]{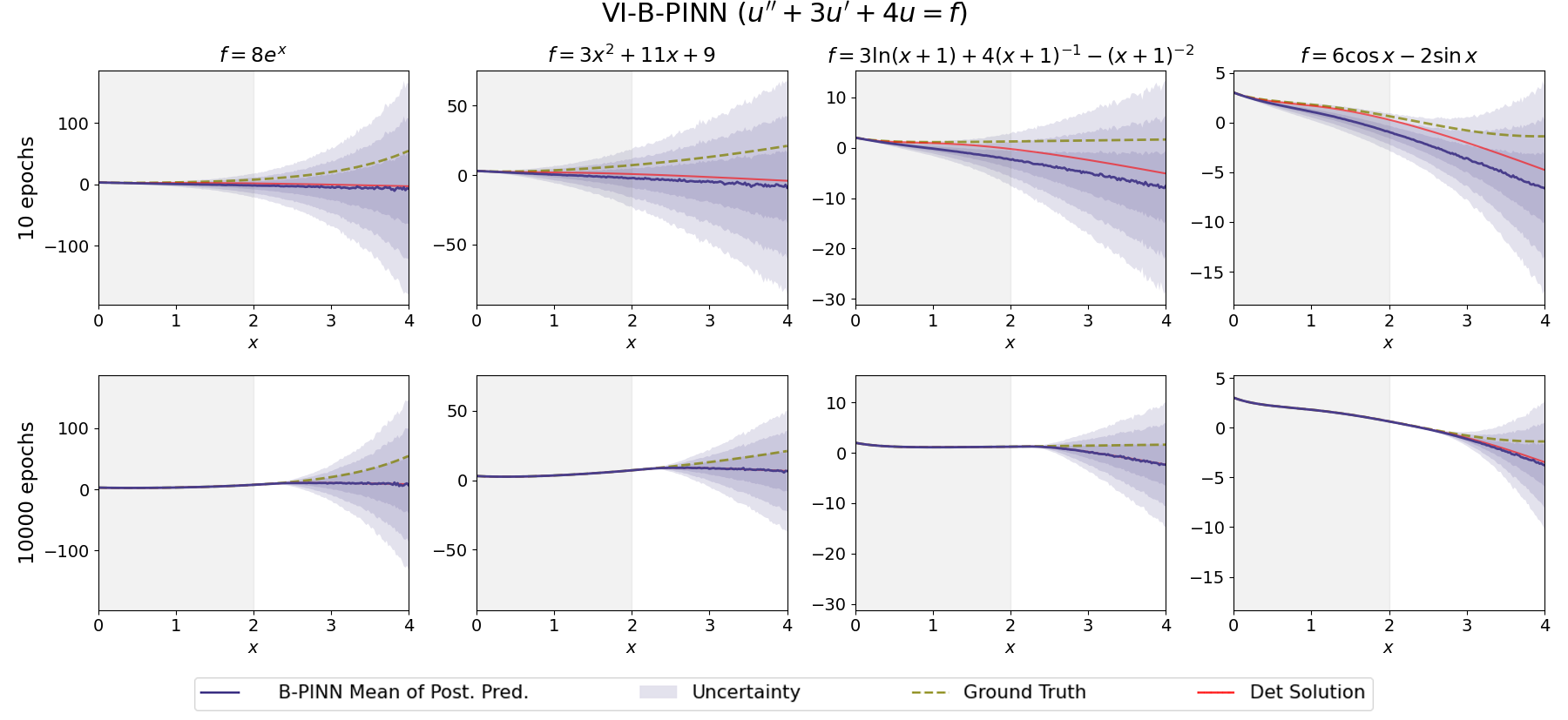}
	\caption{\textbf{Uncertainty estimation with Error-Aware B-PINNs.} We consider equation $u''+3u'+4u=f$ under various source terms. We construct pseudo-aleatoric variance using error bounds, and test our method with Neural Linear Model (NLM-PINN) and Variational Inference (VI-B-PINN).}
	\label{secondorder2}
\end{figure*}

\subsection{Practical Calculation of Error Bound}

We implement error bound (22) in practice by partitioning domain $\Omega$ into subintervals $\Omega=\bigcup_{k=1}^{K}\Omega_k$, $\Omega_k=[n_{k-1},n_{k}]$, and estimating upper bound $\varepsilon_k$ for $r_{\theta}$ on each subinterval, i.e., $\varepsilon_k\geqslant|r_{\theta}(\xi)|$, $\xi\in[n_{k-1},n_{k}]$. The resulting error bound is given by
\begin{align}\label{app1}
	|u(x)-u_{\theta}(x)|\leqslant \lambda^{-1} e^{-\lambda x}\left(\sum_{k=1}^{p}\varepsilon_k\left(e^{\lambda n_k}-e^{\lambda n_{k-1}}\right)+\varepsilon_{p+1}\left(e^{\lambda x}-e^{\lambda n_{p}}\right)\right),\tag{25}
\end{align}
where $p\in\{1,...,K\}$ is chosen such that $n_{p}\leqslant x \leqslant n_{p+1}$.

\subsection{Error-Aware B-PINNs for Second-Order Linear ODEs}
We supplement Section 3.2.1 with examples of a second-order linear ODE (see Figures 4, 5),
\begin{align}\label{app2}
	u''(x)+(\lambda_1+i\omega_1+\lambda_2+i\omega_2) u'(x) + (\lambda_1+i\omega_1)(\lambda_2+i\omega_2)u(x)=f(x).\tag{26}
\end{align}
For the case when $\lambda_1,\lambda_2>0$, the error bound is given by
\begin{align}\label{app3}
	|u(x)-u_{\theta}(x)|\leqslant\frac{1}{\lambda_2- \lambda_1}\Bigg[ &\lambda_1^{-1} e^{-\lambda_1 x}\left(\sum_{k=1}^{p}\varepsilon_k\left(e^{\lambda_1 n_k}-e^{\lambda_1 n_{k-1}}\right)+\varepsilon_{p+1}\left(e^{\lambda_1 x}-e^{\lambda_1 n_{p}}\right)\right)\nonumber\\
	 -&\lambda_2^{-1} e^{-\lambda_2 x}\left(\sum_{k=1}^{p}\varepsilon_k\left(e^{\lambda_2 n_k}-e^{\lambda_2 n_{k-1}}\right)+\varepsilon_{p+1}\left(e^{\lambda_2 x}-e^{\lambda_2 n_{p}}\right)\right)\Bigg].\tag{27}
\end{align}
For the case when $\lambda_1=\lambda_2=0$, it is given by
\begin{align}\label{app4}
	|u(x)-u_{\theta}(x)|\leqslant x\left(\sum_{k=1}^{p}\varepsilon_k\left(n_k - n_{k-1}\right)+\varepsilon_{p+1}\left(x - n_{p}\right)\right)-\left(\sum_{k=1}^{p}\varepsilon_k\left(\frac{n_k^2 - n_{k-1}^2}{2}\right)+\varepsilon_{p+1}\left(\frac{x^2 - n_{p}^2}{2}\right)\right).\tag{28}
\end{align}
Note that the imaginary part $\omega$ and the source term $f$ have no influence on the error estimate. Rigorous proofs regarding the above error bounds can be found in [Liu et al., 2021].

Compared to the first-order case, the uncertainty estimates in Figure 4 and Figure 5 are less tight which is due to the less tight error bound. In case of NLM-PINNs, the prediction mean and the deterministic solution almost coincide. In case of VI-B-PINNs, the prediction mean tends to be less close to the true solution than the deterministic solution, especially if model is insufficiently trained. Regardless of this discrepancy, the uncertainty still covers the true solution.

\section{Experimental Setup}

In this section, we describe the details of our experiments.
\subsection{First-Order and Second-Order Linear ODE Setup}
For all twelve equations we used the same hyperparameters and architecture for the deterministic neural network. We used feed forward neural networks with two hidden layers with 32 units each, hyperbolic tangent as activation function and batch size of 32 points equally spaced within the training domain. The training domain was $[0, 2]$ and we present the results in the testing range $[0, 4]$. We used Adam optimizer with learning rate $0.01$.

The number of epochs vary and they are specified in each figure. We trained the deterministic networks for 10 epochs to evaluate the quality of the uncertainty given by our methods when the networks were undertrained. We also provide the results for 1,000 or 10,000 epochs showing the results when the networks are trained for a sufficient amount of epochs.

To train our NLM-PINN model we discard the output layer of the deterministic network and perform a Bayesian linear regression on top of the feature map generated by the two hidden layers. As we explained in Section 3.2, the setup for NLM-PINN allows for prior optimization. To optimize the standard deviation of the prior we evaluated the objective function on 100 equally spaced standard deviations in the range $[0.1, 1]$.

We used the same architecture and hyperparameters for the VI-B-PINNs. We used a normal distribution 
$\mathcal{N}(0, 0.1)$ and  $\mathcal{N}(0, 1)$ as the prior of the weights for first-order and second-order equations respectively. Since the architecture is shared with the deterministic network we use its weights to initialize the mean of the weights in the VI-B-PINN. This initialization of the weights allows us to freeze the means of the weights which speeds up the training process. To initialize $\rho$ which parameterizes the standard deviation of the weights as $\sigma = log(1 + exp(\rho))$, we sampled from a uniform distribution in the domain $[-5, -4]$. We took 1,000 posterior samples to approximate the posterior predictive. Lastly, all the VI-B-PINNs were trained for 50,000 epochs.

We present the equations and their respective initial conditions in Table 1.

\begin{table}[h]
\caption{First-Order and Second-Order Equations} \label{table:equations}
\begin{center}
\begin{tabular}{lll}
\textbf{EQUATION}  &$u(0)$&$u'(0)$ \\
\hline \\
$u' + 3u = 3t^2+5t+4$         & 2.0 & - \\
$u' + 3u = 6\cos{3t}$         & 2.0 & - \\
$u' + 3u = 4e^t$         & 2.0 & - \\
$u' + 3u = -9\mathsf{ln}(t+1)-(1-t)^{-2}$         & 2.0 & - \\\\
\hline \\
$u'' + u = 2e^t$         & 2.0 & 2.0 \\
$u'' + u = t^2+t+3$             & 2.0 & 2.0 \\
$u'' + u = \mathsf{ln}(t+1)-(t+1)^{-2}$             & 1.0 & 2.0 \\
$u'' + u = 2\cos{t^2}+(1-4t^2)\sin{t^2}$& 1.0 & 1.0 \\
$u'' + 3u' + 4u = 8e^t $ & 3.0 & -3.0\\
$u'' + 3u' + 4u = 3t^2+11t+9 $ & 3.0 & -3.0\\
$u'' + 3u' + 4u = 3\mathsf{ln}(t+1)+4(t+1)^{-1}-(t+1)^{-2} $& 2.0 & -3.0 \\
$u'' + 3u' + 4u = 6\cos{t}-2\sin{t} $& 3.0 & -3.0\\
\end{tabular}
\end{center}
\end{table}

\subsection{Burgers' PDE Setup}
Similarly to the previous setup, for Burgers' equation we also use a feed forward neural network with two hidden layers, 32 units on each of them and sigmoid activation. We optimized using Adam with learning rate $10^{-3}$ for 20,000 epochs.

We trained the networks in the domain $\{(x,t) | x \in [-1, 1], t \in [0,1]\}$ and we show the results in the testing domain $\{(x,t) | x \in [-1, 1], t \in [0,2]\}$ where $x$ is the spatial coordinate and $t$ the temporal coordinate. To train the network we take a grid of $100\times100$ equally spaced points in the training domain and add noise to them. We used the equation parameter $\nu=\frac{0.01}{\pi}$.

The VI-B-PINN model was trained with the same architecture and hyperparameters. We used a normal distribution $\mathcal{N}(0, 1)$ as the prior of the weights. Since there are no error bounds available for PDEs we use residuals to define the operator needed in (13). One approach would be to simply take the residual $r_{MSE}(x, t)$. Instead we take the accumulated residuals through time as:

$$\mathcal{E}[r_{\mathsf{MSE}}(x, t), \lambda(x, t)] = \int_{t_0}^{t}r_{\mathsf{MSE}}(x, \tau)d\tau\approx(t - t_0)\frac{1}{N}\sum_{i=0}^N r_{\mathsf{MSE}(x, t_i)},$$

where $t_i$ are equally spaced.

We set the following initial and boundary conditions:
\begin{align*}
    u(x=-1, \tau)&=0 \; \forall \tau \in [0, 1],\\
    u(x=1, \tau)&=0 \; \forall \tau \in [0, 1],\\
    u(x', t=0)&=-\sin{\pi x'}.\\
\end{align*}

\subsection{Existing B-PINN Approach}
We solved first-order equations using the existing B-PINN approach, the results are shown in Figure 1. In this case we place the likelihood over the residuals and use a homoscedastic variance as described in Section 3.1. For this experiment we used the same setup described before for first-order equations. As homoscedastic variance we chose $\sigma_{\mathcal{D}}^2=1$.
\vfill

\end{alphasection}